# Detecting Concept-level Emotion Cause in Microblogging

Shuangyong Song, Yao Meng
Internet Application Laboratory, Fujitsu R&D Center Co., Ltd.
Beijing 100025, China.
{shuangyong.song, mengyao}@cn.fujitsu.com

## ABSTRACT
In this paper, we propose a Concept-level Emotion Cause Model (*CECM*), instead of the mere word-level models, to discover causes of microblogging users' diversified emotions on specific hot event. A modified topic-supervised biterm topic model is utilized in *CECM* to detect 'emotion topics' in event-related tweets, and then context-sensitive topical PageRank is utilized to detect meaningful multiword expressions as emotion causes. Experimental results on a dataset from *Sina Weibo*, one of the largest microblogging websites in China, show *CECM* can better detect emotion causes than baseline methods.

**Categories and Subject Descriptors:** H.2.8 [**Database Management**]: Database Applications – Data mining.

## 1. INTRODUCTION
Fast diffusion of information makes microblogging a convenient platform for users to seek new trends and express emotions about hot events. Generally, users may have different emotions on an event, and those emotions are from different aspects of the event with various causes [1]. For example, with event 'missing Malaysia plane MH370', victim relatives' hearts are filled with sadness and fear, while other people may feel sympathetic and angry. By analyzing people's emotions and the causes, a user can better understand the details of events he is interested in, such as events' different perspectives and development tendency.

Existing studies have focused on detecting emotion cause by discovering the most frequently co-occurred clauses or words with a single emotion word [1]. Those approaches are beneficial when the average length of texts is not short. However, it is hard to discover the co-occurrence relationship between emotions and their causes in tweet-like short texts. Furthermore, causes of a single emotion word can't thoroughly describe information of a specific emotion, since emotion is an abstract topic. To solve those problems, in this paper we proposed a Concept-level Emotion Cause Model (*CECM*) for detecting emotion topics and emotion causes in microblogging.

Microblogging users may use various expression to express their emotions, more than just use standard emotion words [3]. In addition, those non-emotion-word expression on different events are also generally not same [6]. So taking emotion words as seeds to automatically detect user emotion on specific event is more labor-saving and robust than training model with massive manual annotations. *CECM* first preprocesses the microblogging data and perform Chinese word segmentation, and then utilizes a topic-supervised biterm topic model to detect emotion topics with a manually collected emotion dictionary, finally detects emotion causes from the emotion topics, and it can also detect relationship between emoticons and emotions. In the following of this article, we will illustrate the details of problem definition and proposed *CECM* model, along with the experimental results.

## 2. PROBLEM DEFINITION & METHOD
### 2.1 Problem Definition
Referring to the description in [1], we give the definition of *emotion topic* as 'a probability distribution over words according to their relevance to a specific emotion', and assume that emotion causes are contained in related emotion topic. In our emotion dictionary, number of emotions is $K$, and emotions are defined as $E = \{e_1, e_2, …, e_K\}$. The number of emotion topics is also $K$, decided by emotion dictionary, and emotion topics are accordingly denoted as $T_e = \{t_{e1}, t_{e2}, …, t_{eK}\}$. Assuming the number of emoticons in our data is $N$, we define emoticons as $I = \{i_1, i_2, …, i_N\}$. Given an event related microblogging dataset, our goal is to detect causes of concept-level emotions from $T_e$.

### 2.2 Method Description
*CECM* contains four modules: preprocessing and word segmentation, emotion topic detection, emotion cause detection and emoticon-emotion relationship detection.

**Preprocessing and Word Segmentation** The "@username" symbols are removed since they does not actually represent meaningful content or emotions. URLs, non-texts and 'forwarding' tags are also removed. Besides, microblogging allows users to insert emoticons in tweets to express sentiments, such as '[anger]', '[yeah]' and '[kiss]'. For detecting relation between emoticons and users' emotions, we keep those emoticons as simple terms. Then we apply ICTCLAS system [2] to perform Chinese word segmentation on the corpus, with taking all emotion words in our collected emotion dictionary as user defined words. Then tweets containing at least one emotion word will be remained in final experimental data.

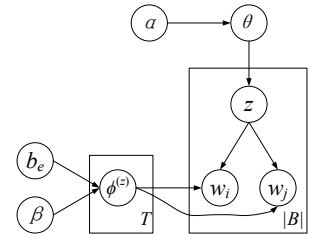

**Fig. 1. Topic-supervised biterm topic model**

**Emotion Topic Detection** The graphical representation of proposed topic-supervised biterm topic model (*TS-BTM*) is given in Fig. 1, where $\phi$ represents the distribution of words on $T_e$ and $\theta$ represents the distribution of $T_e$ on tweets. $\alpha$ and $\beta$ are hyperparameters of Dirichlet priors for $\theta$ and $\phi$, respectively. $|B|$ is the total number of biterms [4]. For discovering $T_e$, we add binary distributions $b_e$ of emotion words on different emotions into the biterm topic model, which means that the probability of an emotion word appearing in the corresponding emotion topic is 1 and in other emotion topics is 0, to restrict each topic can just describe one emotion.

**Emotion Cause Detection** With the *TS-BTM*, we can get $T_e$. Then we can detect causes for each emotion $e_k$ from $t_{ek}$ with context-sensitive topical PageRank (*cTPR*) method proposed in [5]. *cTPR* takes terms as nodes in term graph, and the weight of edge from term $x$ to term $y$ is decided by the time of y showing as a previous term of x in topical tweets, then get topical ranking value of each term. After term ranking using *cTPR*, a common keyphrase generation method is used to detect keyphrases in each $t_{ek}$ as emotion cause. We first select top 100 non-emotional terms for each $t_{ek}$, and then look for combinations of these terms that occur as frequent phrases in the text collection [6].

**Emoticon-Emotion Relation Detection** We define $R(e_k, i_n)$ as relevance value between $e_k$ and $i_n$, which is contained in the detection result of $t_{ek}$. We set a *threshold* in formula (1) to judge if an emoticon is possibly related to an emotion, where $p$ is a factor for adjusting the threshold value. Then, we utilize the ranking of remained *threshold*-exceeded emoticons in each emotion topic to judge the importance of them.

$$threshold = p * \frac{\sum_{k=0}^{K}\sum_{n=0}^{N} R(e_k, i_n)}{K*N} \quad (1)$$

## 3. EXPERIMENTS

### 3.1 Dataset and Parameter Setting
*Sina Weibo* is a well-known Chinese micro-blogging service, and the dataset used in our experiments is collected from *Sina Weibo*, which is about the 2011 Japanese Earthquake. It contains 42,109 tweets, comprised of 17,929 different words and 828 emoticons. Emotion dictionary used in our experiments is collected manually with considering some different related resources, which contains 36 emotions and accordingly 542 emotion words. Table 1 gives some examples. We set the parameter $K$ to be 36 according to this dictionary and $N$ to be 828 according to the microblogging data.

**Table 1. Emotion dictionary examples**

| Emotions | Chinese Emotion words |
|---|---|
| Esteem | 敬佩(jingpei), 敬仰(jingyang), 敬重(jingzhong) |
| Sadness | 哭(ku), 悲痛(beitong), 叹息(tanxi), 悲哀(beiai) |
| Sympathy | 同情(tongqing), 可怜(kelian), 可惜(kexi) |

### 3.2 Results and Discussions
We measure the effects of *CECM* using Normalized Discounted Cumulative Gain at top k (*NDCG@k*) and Mean Average Precision (*MAP*). Accordingly, two additional emotion cause detection methods are conducted as baselines:
**Co-occurrence method**: using the co-occurrence method given in [1] to detect emotion causes.
**Uniterm LDA method:** detecting concept-level emotion causes but using a uniterm topic model to detect emotion topics.

For experimental evaluation, we have 3 tagging volunteers manually label the emotion causes. Since users are always just care about the most important causes, we just evaluate top 10 results of each method on each emotion. After deduping of those causes, each annotator should give them integral scores between 1 and 5, and finally we calculate the average. Ranking result of them will be used to get *NDCG@k* value and 10 causes with maximum scores are set to be the 'right causes' to get *MAP* value.

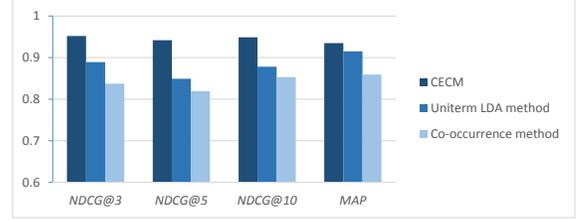

**Fig. 2. Evaluation results of comparing three methods**

Fig. 2 demonstrates the evaluation results, from which we can see *CECM* outperforms *Co-occurrence method* and *Uniterm topic method*. The improvement is due to the effect of biterm topic model on analyzing short texts. In addition, two concept-level methods perform better than co-occurrence based mere word-level method, this result proves that emotion is an abstract concept and detection of implicit emotion expressions is important. Table 2 gives some emotion cause detection result examples. We can see the most possible reasons for users' esteem emotion include 'nuclear power plant workers', 'Japan self-defense force' and 'rescue team', since they appeared calm and brave after the earthquake and nuclear leakage.

**Table 2. Emotion cause detection result examples**

| Emotions | Emotion causes |
|---|---|
| Esteem | nuclear power plant workers, self-defense force, rescue team |
| Sadness | adopt orphans after the earthquake, the Libyan war |
| Sympathy | nearby residents, evacuation, nuclear radiation |

Table 3 shows some 'emoticon-emotion' relationship examples. Some surprising results exist such as that we may empirically think [scold] and [anger] can just express anger emotion, actually, in special events they can also express intense sadness. Based on manual observation, we roughly set *p* in formula (1) as 0.6.

**Table 3. 'Emotion-Emoticon' relationship examples**

| Emotions | Emoticons |
|---|---|
| Esteem | [lovely], [fist], [breeze], [jostle], [cult] |
| Sadness | [scold], [anger], [shocked], [wave], [orz] |
| Sympathy | [sad], [candle], [handshaking], [scared], [sick] |

## 4. CONCLUSIONS AND FUTURE WORK
To the best of our knowledge, no other concept-level emotion cause detection work has been done to date. In future work, we will design a new concept-level emotion analysis model for key emotion detection from event related micro-blogging corpus, and emotion cause tracking are also part of our future work.

## 5. REFERENCES

[1] Lee, S., Chen, Y., Huang, C-R., Li, S-S. Detecting emotion causes with a linguistic rule-based approach. Computational Intelligence.
[2] Zhang, H., Yu, H., Xiong, D., Liu, Q. HHMM-based Chinese lexical analyzer ICTCLAS. In second SIGHAN workshop on Chinese language processing, pp 184-187.
[3] Li, D., Shuai, X., Sun, G., Tang, J., Ding, Y., Luo, Z. Mining topic-level opinion influence in microblog. In CIKM'12.
[4] Yan, X., Guo, J., Lan, Y., Cheng, X. 2013. A biterm topic model for short texts. In WWW'13, pp 1445–1456.
[5] Zhao, W., Jiang, J., He, J., Song, Y., Achananuparp, P., Lim, E-P., Li, X. Topical keyphrase extraction from twitter. In ACL'11.
[6] Song, S., Meng, Y., Sun, J. Detecting keyphrases in microblogging with graph modeling of information diffusion, PRICAI'14, pp 26-38.